\documentclass{article} 
\usepackage{colm2024_conference}
\usepackage[T1]{fontenc}
\usepackage{microtype}
\usepackage{hyperref}
\usepackage{url}
\usepackage{booktabs}
\usepackage{CJKutf8}
\usepackage{graphicx}
\usepackage{multicol}
\usepackage{multirow}
\usepackage{array}
\usepackage{amsmath}
\usepackage{booktabs}
\usepackage{enumitem}
\usepackage{wrapfig}
\usepackage{algorithm}
\usepackage{algpseudocode}
\usepackage{microtype}
\usepackage{amsmath}
\usepackage{colortbl}
\usepackage[utf8]{inputenc}
\usepackage{amssymb}
\definecolor{lightgray}{rgb}{0.9,0.9,0.9}
\definecolor{rowgray}{gray}{0.95}  
\usepackage{caption}
\usepackage{subcaption}
\usepackage{xcolor}
\usepackage{setspace}
\usepackage{url}
\usepackage{colortbl}
\usepackage{tabularx}
\usepackage{blindtext}
\usepackage{pgfplots}
\pgfplotsset{compat=1.18} 
\usepackage{tikz}
\usetikzlibrary{er,positioning,bayesnet}
\usepackage{makecell}
\usepackage{tipa}
\usepackage{siunitx}
\usepackage{nicefrac}
\usepackage{tocloft}
\usepackage{listings}
\usepackage{tcolorbox}
\usepackage{xltabular}
\usepackage{adjustbox}
\usepackage{xurl}
\usepackage{setspace}
\usepackage{lipsum}
\usepackage{longtable}

\title{OmniAlign: A Unified Multilingual Aligner for Word and Sentence Alignment}

\author{
	Mengpeng Yang \ \ \ \ \ Jingxu Yang \ \ \ \ \ Chao Chen \ \ \ \ \ Tian Xia \ \ \ \ \ Yabo Sun \ \ \ \ \ Qiang Liu \\
	WPS Qingqiu / Wuhan, China \\
	\texttt{yangmengpeng@wps.cn, yangjingxu1@wps.cn, chenchao9@wps.cn} \\
	\texttt{xiatian3@wps.cn, sunyabo@wps.cn, liuqiang2@wps.cn}
}
\begin{document}
\begin{CJK*}{UTF8}{gbsn}
\maketitle
		
\begin{abstract}
Cross-lingual sequence alignment is fundamental for building and exploiting parallel corpora, spanning mappings from documents and sentences down to words and subwords. Existing tools, however, typically specialize in a single granularity, so practitioners often need separate systems for word- and sentence-level alignment---especially in multilingual and long-text settings. We present OmniAlign, a unified multilingual aligner that supports both word-level and sentence-level alignment with a single lightweight model. Built on an encoder-only backbone with strong long-context modeling, OmniAlign induces word alignments from contextualized token similarity matrices, and obtains document-level $m$--$n$ sentence alignments via sentence embeddings combined with dynamic programming. To balance fine-grained alignment accuracy and sentence-representation quality, we use a four-stage training pipeline: alignment-oriented continued pre-training, self-supervised learning, supervised fine-tuning on human annotations, and sentence-embedding distillation from a strong multilingual teacher. Experiments show that OmniAlign achieves highly competitive performance on both word- and sentence-alignment benchmarks and generalizes well to unseen language pairs. Surprisingly, later-stage supervised fine-tuning on short texts further improves alignment quality while retaining the long-context understanding acquired in earlier training, keeping the model robust on long-text word alignment.

\normalsize {\color{blue}\textbf{Code}: \url{https://github.com/MilkDargon/OmniAlign}}\par
{\color{blue}\textbf{Model}: \url{https://huggingface.co/WPS-Qingqiu/OmniAlign}}
\end{abstract}

\section{Introduction}
\textbf{Cross-lingual Sequence Alignment} is a core foundational task in the construction and exploitation of parallel corpora. Its goal is to establish precise and consistent correspondences between text sequences across different languages. This task typically spans multiple levels of granularity—from coarse-grained alignment at the document or paragraph level, to sentence-level alignment, and further down to fine-grained word- or subword-level alignment. Together, these alignment processes form the critical pathway for cross-lingual information mapping, enabling models to build stable semantic correspondences in multilingual spaces. They provide essential support for tasks such as building para-crawl pipelines, constructing translation memories in computer-assisted translation systems \citep{varga2008parallel}, projecting linguistic annotations and XML structural labels across languages \citep{david2001inducing, hashimoto2019high}, detecting under-translation in neural machine translation \citep{tu2016modeling}, developing cross-lingual transfer NLP tools \citep{nicolai2019learning, mayhew2017cheap}, and improving or augmenting multilingual pre-trained models \citep{chi2021improving}.

\begin{figure}[htbp]
    \centering
    \begin{subfigure}[b]{0.40\linewidth}
        \centering
        \includegraphics[width=\linewidth]{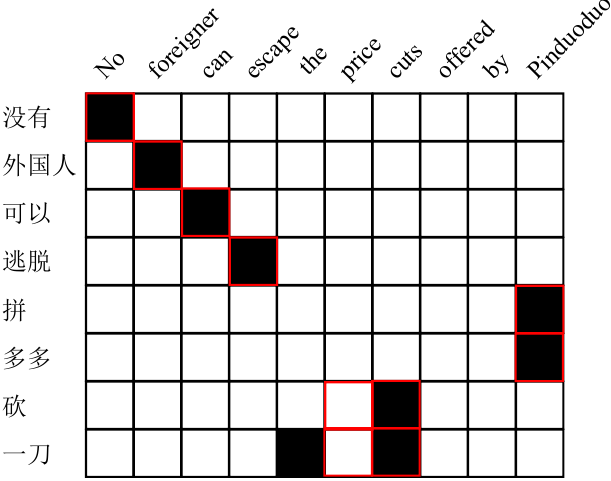}
        \caption{Word Alignment Example. Red boxes denote the gold alignments.}
        \label{fig:1_task_a}
    \end{subfigure}
    \hfill
    \begin{subfigure}[b]{0.58\linewidth}
        \centering
        \includegraphics[width=\linewidth]{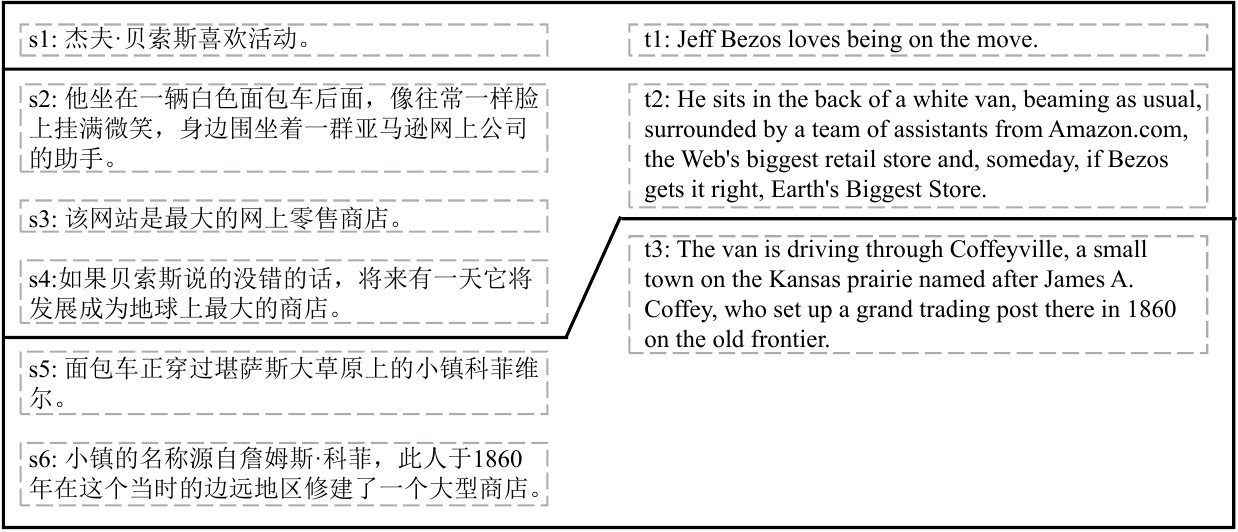}
        \caption{Sentence Alignment Example.}
        \label{fig:1_task_b}
    \end{subfigure}
    \caption{Sequence Alignment Diagram.}
    \label{fig:1_task}
\end{figure}

Despite the emergence of various models targeting sequence alignment at different granularities in recent years, there is still a lack of a single practical aligner that covers both word- and sentence-level needs while remaining multilingual and long-context capable. Previous studies typically focus on only one subtask of the alignment problem. For example, methods such as \citep{sabet2020simalign, dou2021word, wang2022multilingual, wu2023wspalign, nagata2020supervised, lai2022cross, latouche2024binaryalign} primarily explore word alignment; however, most of them suffer from significant performance degradation as sequence length increases and are unable to handle alignment scenarios exceeding 512 tokens. Some approaches claim multilingual support, yet they require training separate models for different languages and do not also provide sentence-level alignment in the same model.

On the other hand, methods such as \citep{molfese2024crocoalign, steingrimsson2023sentalign, thompson2019vecalign, liu2023bertalign} rely on cross-lingual representation models like LaBSE \citep{feng2022language} or LASER \citep{artetxe2019massively} to obtain sentence embeddings and determine sentence alignment via similarity computation. However, these methods do not provide fine-grained word-level alignment information.

To address these limitations, we propose OmniAlign—a unified multilingual aligner that supports both word-level and sentence-level alignment with a single model. Word and sentence outputs are obtained from the same encoder via task-specific alignment procedures, enabling developers to meet diverse cross-lingual alignment needs without maintaining separate systems. The main contributions of this work are as follows:
\begin{enumerate}[left=2pt, itemsep=2pt, topsep=4pt]
	\item \textbf{We introduce OmniAlign}. A novel open-source multilingual sequence alignment model with only 0.3B parameters and efficient inference. It supports bidirectional word- and sentence-level alignment across eleven major languages.
	\item \textbf{Multi-stage training recipe}. We design a multi-stage training framework that first enhances the model’s token-level semantic representations through continued pre-training. We then combine self-supervised learning with supervised signals to optimize token-level alignment accuracy and obtain the word-alignment model. Finally, we freeze the word-alignment module and employ knowledge distillation to restore and further strengthen the model’s sentence-level representation capability.
	\item \textbf{Broad applicability and competitive performance}. OmniAlign supports a wide range of alignment tasks encountered in real-world applications and achieves highly competitive results on both sentence-level and word-level alignment benchmarks, remaining robust on long-text word alignment—where short-text supervised fine-tuning further improves quality—and generalizing well to unseen language pairs.
\end{enumerate}
\begin{table}[h]
\centering
\small
\label{tab:alignment_methods_simple}
\begin{tabular}{lcccc}
\toprule
\textbf{Methods} & \textbf{Multilingual} & \textbf{Long Text} & \textbf{Word Alignment} & \textbf{Sentence Alignment} \\
\hline
\rowcolor{rowgray}
FastAlign \citep{dyer2013simple}        &  &  & \checkmark &  \\

GIZA++ \citep{och2003systematic}        &  &  & \checkmark &  \\

\rowcolor{rowgray}
SimAlign \citep{sabet2020simalign}        & \checkmark &  & \checkmark &  \\

AwesomeAlign \citep{dou2021word}    & \checkmark &  & \checkmark &  \\

\rowcolor{rowgray}
AccAlign \citep{wang2022multilingual}        & \checkmark &  & \checkmark &  \\

WSPAlign \citep{wu2023wspalign}        & \checkmark &  & \checkmark &  \\

\rowcolor{rowgray}
BinaryAlign \citep{latouche2024binaryalign}     &  &  & \checkmark &  \\

Gale-Church \citep{gale1993program}     &  & \checkmark &  & \checkmark \\

\rowcolor{rowgray}
BleuAlign \citep{sennrich2011iterative}      & \checkmark & - &  & \checkmark \\

VecAlign \citep{thompson2019vecalign}        & \checkmark & - &  & \checkmark \\

\rowcolor{rowgray}
BertAlign \citep{liu2023bertalign}       & \checkmark & - &  & \checkmark \\

SentAlign \citep{steingrimsson2023sentalign}       & \checkmark & - &  & \checkmark \\

\rowcolor{rowgray}
CrocoAlign \citep{molfese2024crocoalign}      & \checkmark & - &  & \checkmark \\

\hline

\textbf{OmniAlign} & \checkmark & \checkmark & \checkmark & \checkmark \\
\bottomrule
\end{tabular}
\caption{This report compares OmniAlign with a range of existing state-of-the-art sequence alignment methods. The symbol “\textbf{–}” indicates that the capability is independent of the specific model choice.}
\end{table}
The remainder of this paper is organized as follows: we first describe the training methodology of OmniAlign, then present experimental results for each stage of the model variants, and finally conclude with key findings.

\section{OmniAlign}
This section focuses on the operational mechanism of OmniAlign in sequence alignment tasks and elaborates on its training pipeline.

\begin{figure}[htbp]
\centering
\includegraphics[width=1\linewidth]{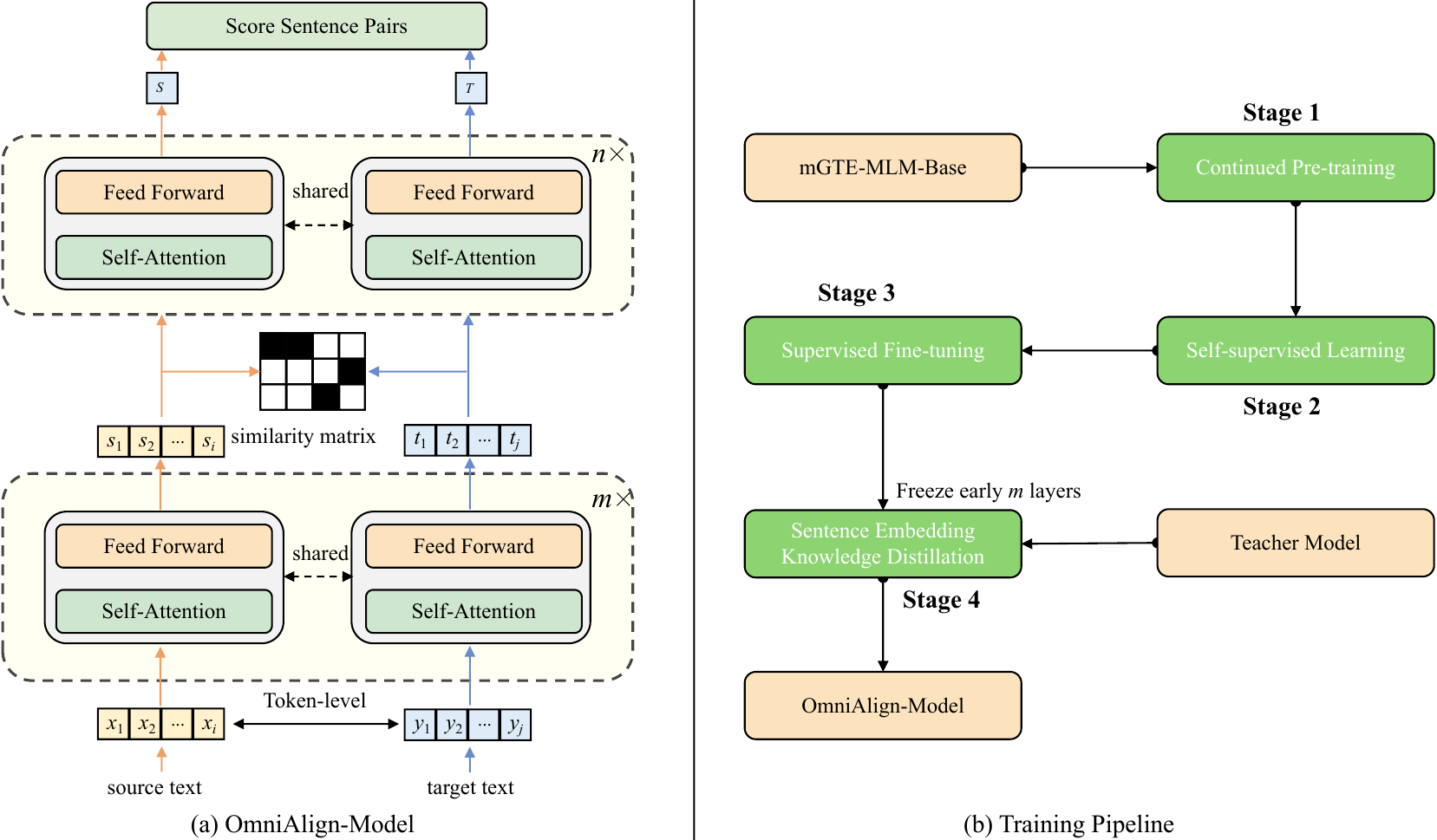}
\caption{Overall architecture of OmniAlign, including the sequence alignment model (left) and the multi-stage training pipeline (right).}
\label{fig:OmniAlign_and_its_training_pipeline}
\end{figure}

\subsection{Base Model}
We adopt the mGTE base model \citep{zhang2024mgte} as the base model for OmniAlign, in order to support multi-granularity, cross-lingual, and variable-length sequence alignment. The main reasons are as follows:

\begin{itemize}[left=2pt, itemsep=2pt, topsep=4pt]
    \item \textbf{Encoder-only architectures provide stronger contextual modeling.} 
    Compared with Seq2Seq or decoder-only models, encoder-only models offer inherent advantages in capturing intra-sentence token dependencies. This enables more fine-grained semantic representation learning, making them particularly suitable for word-level alignment tasks.

    \item \textbf{Cross-lingual alignment requires robust multilingual capability.} 
    Effective word- and sentence-level alignment across languages requires the model to have strong and reliable multilingual capability.

    \item \textbf{mGTE overcomes the sequence-length limitations of traditional multilingual models.} 
    Conventional models such as mBERT \citep{pires2019multilingual} and the XLM family \citep{conneau2020unsupervised} are constrained by a maximum sequence length of 512 tokens, making them insufficient for long-text alignment scenarios. In contrast, mGTE supports significantly longer input sequences, enabling alignment over extended contexts.
\end{itemize}

\subsection{Extracting Word Alignments from Token Embeddings}
To enable OmniAlign to support both word-level and sentence-level alignment, we induce word alignments from the contextualized token embeddings of multilingual pre-trained language models.

As shown in Figure~\ref{fig:OmniAlign_and_its_training_pipeline}(a), given a source text $\mathbf{x} = \langle x_{1}, x_{2}, \ldots, x_{i} \rangle$ and a target text $\mathbf{y} = \langle y_{1}, y_{2}, \ldots, y_{j} \rangle$, we denote the contextualized token embeddings from the $m$-th layer of OmniAlign as $\mathbf{s} = \langle s_{1}, s_{2}, \ldots, s_{i} \rangle$ and $\mathbf{t} = \langle t_{1}, t_{2}, \ldots, t_{j} \rangle$, respectively. Following prior work \citep{sabet2020simalign, dou2021word, wang2022multilingual}, we compute the dot product between $\mathbf{s}$ and $\mathbf{t}$ to obtain the similarity matrix.
\begin{equation}
    \mathbf{S} = \mathbf{s}\mathbf{t}^\mathrm{T},
\end{equation}
each row of the similarity matrix $\mathbf{S}$ is normalized using a Softmax function to obtain the source-to-target probability alignment matrix $\mathbf{S}_{xy}$. The i-th row of $\mathbf{S}_{xy}$ represents the alignment probabilities between token $x_i$ and all tokens in $\mathbf{y}$. Similarly, we compute the target-to-source probability alignment matrix $\mathbf{S}_{yx}$, and the final alignment matrix is obtained by taking the intersection of the two probability matrices.
\begin{equation}
    \mathbf{A} = (\mathbf{S}_{xy} > c) * (\mathbf{S}_{yx}^T > c),
\end{equation}
where $c$ denotes the threshold, and $\mathbf{A}_{ij} = 1$ indicates that $x_i$ and $y_j$ are aligned. The above procedure induces alignments at the token level; if any pair of tokens is aligned, the corresponding words are aligned to produce the final word-level alignment results.

\subsection{Extracting Sentence Alignments From Sentence Embeddings}
To support sentence-level parallel corpus alignment, OmniAlign integrates a two-stage alignment algorithm inspired by Bertalign \citep{liu2023bertalign}.

\textbf{Similarity Retrieval-Based 1–1 Alignment.} Given a source document $\mathbf{X} = \langle X_{1}, X_{2}, \ldots, X_{N} \rangle$ and a target document $\mathbf{Y} = \langle Y_{1}, Y_{2}, \ldots, Y_{M} \rangle$, sentence encoders map each sentence to a dense vector representation, yielding source sentence embeddings $\mathbf{S} = \langle S_{1}, S_{2}, \ldots, S_{N} \rangle$ and target sentence embeddings $\mathbf{T} = \langle T_{1}, T_{2}, \ldots, T_{M} \rangle$.

We first perform maximum inner-product search to retrieve the top-$k$ candidate target sentences for each source sentence:
\begin{equation}
    (\mathbf{D}, \mathbf{I}) = \mathrm{TopK}\big(\mathbf{S}\mathbf{T}^{\mathrm{T}}, k\big),
\end{equation}
where $\mathbf{S}\mathbf{T}^{\mathrm{T}}$ denotes the similarity matrix between source and target sentences. The top-$k$ operation yields the similarity score matrix $\mathbf{D}$ and the corresponding target sentence index matrix $\mathbf{I}$ for each source sentence.

Subsequently, a dynamically adjusted diagonal search window is constructed, within which dynamic programming is applied to determine the optimal alignment path. The state transition $(a_1, a_2) \in \{(0,1), (1,0), (1,1)\}$ corresponds to target sentence insertion, source sentence deletion, and one-to-one sentence alignment, respectively. The scoring function is defined as:
\begin{equation}
    \mathrm{score}(i,j)=\max_{(a_1,a_2)}
    \left[
    \mathrm{score}(i-a_1,j-a_2) + \delta_{1-1}(a_1,a_2,i,j)
    \right].
\end{equation}
Here, $\delta_{1-1}$ is an indicator function that contributes the similarity score from $\mathbf{D}$ only when the alignment operation is (1,1) and the target sentence $j$ belongs to the top-$k$ candidate set $\mathbf{I}$ of the source sentence $i$; otherwise, it contributes zero.

Finally, by backtracking the dynamic programming table, we obtain a sequence of one-to-one alignment anchor points for the entire document.

\textbf{Anchor-Constrained $m$--$n$ Alignment.} Guided by the 1–1 anchors, we restrict the second-stage search to local regions around semantically adjacent anchor points, enabling flexible $m$–$n$ alignments. In the second stage, more flexible alignment patterns are permitted $(a_1, a_2)$ (where $a_1+a_2 \ge 2$). The alignment score for this stage is defined as:
\begin{equation}
    \mathrm{sim}(i,j,a_1,a_2)=
    \langle \mathbf{S}_{a_1}(i), \mathbf{T}_{a_2}(j) \rangle \times \mathrm{LengthPenalty},
\end{equation}
where, $\langle \mathbf{S}_{a_1}(i), \mathbf{T}_{a_2}(j) \rangle$ denotes the inner product between the embedding vectors of the source text span $\mathbf{S}_{a_1}(i)$ and the target text span $\mathbf{T}_{a_2}(j)$, while $\mathrm{LengthPenalty}$ represents a penalty term based on the token-length ratio difference. Compared to constraints based on sentence counts, this token-level ratio is more robust in scenarios involving languages with substantial structural differences, such as Chinese and English. 

The final alignment results are obtained by backtracking the dynamic programming table in the second stage, and can be represented as: $\mathcal{A}=\{([i_1{:}i_2], [j_1{:}j_2]),\ldots\}.$ $\mathcal{A}$ denotes the document-level $m-n$ alignment set. The terms $[i_1{:}i_2], [j_1{:}j_2]$ correspond to contiguous segments in the source document spanning from the $i_1$-th sentence to the $i_2$-th sentence, and contiguous segments in the target document spanning from the $j_1$-th sentence to the $j_2$-th sentence, respectively. Each pair of $([i_1{:}i_2], [j_1{:}j_2])$ constitutes a minimal sentence-level alignment unit.
\begin{table}[htbp]
\centering
\caption{Supported Languages Evaluated}
\label{tab:Supported_Languages}
\begin{tabular}{cccccc}
\hline
\textbf{Languages} & \textbf{ISO 639-1} & \textbf{Languages} & \textbf{ISO 639-1} & \textbf{Languages} & \textbf{ISO 639-1} \\ \hline
Chinese            & zh                 & Spanish            & es                 & Italian            & it                 \\
English            & en                 & Japanese           & ja                 & Korean             & ko                 \\
French             & fr                 & Russian            & ru                 & German             & de                 \\
Portuguese         & pt                 & Romanian           & ro                 &                    &                    \\ \hline
\end{tabular}
\end{table}

\subsection{Training Pipeline}
To enable OmniAlign to achieve strong performance in both word-level and sentence-level alignment, we design a four-stage training strategy (as illustrated in Fig.~\ref{fig:OmniAlign_and_its_training_pipeline}(b)).

\textbf{Continued Pre-training.} We first evaluate the initial mGTE base model on word alignment (see Section \ref{subsec:3.5.2}). The results show that its zero-shot alignment capability is relatively limited. Following the previous works \citep{pires2019multilingual,  conneau2020unsupervised, gururangan2020don, conneau2019cross}, we further conduct continued pre-training on mGTE.

Specifically, given a parallel sentence pair $(x, y)$, we construct four input variants: $x, y, [x,y]$, and $[y,x]$. From each input, 15\% of tokens are randomly selected as prediction targets. Among the selected tokens, 80\% are replaced with the \texttt{[MASK]} token, 10\% are left unchanged, and the remaining 10\% are randomly substituted with other tokens from the vocabulary. The model is then trained to recover the original masked tokens based on the surrounding context.
\begin{equation}
    \mathcal{L}_{\text{1}} = \log p(x\mid x^{\text{mask}})+\log p(y \mid y^{\text{mask}})+\log p(x,y \mid [x,y]^{\text{mask}})+\log p(y,x \mid [y,x]^{\text{mask}}).
\end{equation}
The source texts for continued pre-training are drawn from FineWeb2\footnote{\url{https://huggingface.co/datasets/HuggingFaceFW/fineweb-2}}, covering a mixture of short and long documents. Corresponding translations are synthesized using Qwen3-235B-A22B\footnote{\url{https://huggingface.co/Qwen/Qwen3-235B-A22B}} to construct a parallel corpus of approximately 9 million pairs.

\textbf{Self-supervised Learning.} Let $\hat{\mathbf{A}}$ denote the alignment labels for a given sentence pair $\mathbf{x}$ and $\mathbf{y}$. During training, we derive pseudo-labels $\hat{\mathbf{A}}$ from the model’s online predictions of alignment relations and use them to construct the training objective. For this stage, we sample 100,000 pairs from the parallel corpus constructed during continued pre-training, with an emphasis on long-text examples, for self-supervised optimization.

\begin{equation}
    \mathcal{L}_2 = \sum_{ij} \hat{\mathbf{A}}_{ij} \frac{1}{2} \left( \frac{(\mathbf{S}_{xy})_{ij}}{n} + \frac{(\mathbf{S}_{yx}^{\text{T}})_{ij}}{m}\right).
\end{equation}
This objective encourages the model to bring the contextual representations of aligned words closer together, while additionally enforcing a consistency constraint between $\mathbf{S}_{xy}$ and $\mathbf{S}_{yx}$. Such a constraint promotes symmetry between the alignment probability matrices in both directions, thereby improving alignment quality and training stability.

\textbf{Supervised Fine-tuning.} In the final stage of word alignment training, the training objective remains consistent with that of the self-supervised learning stage, while the dataset used in this stage is described in Section~\ref{subsec:datasets}. The key difference lies in performing supervised fine-tuning with human-annotated gold alignment labels instead of pseudo-labels, thereby further improving the model’s alignment accuracy.

\textbf{Sentence Embedding Knowledge Distillation.}
To rapidly equip our trained word alignment model with sentence-level representation capability, we adopt a knowledge distillation approach following \citep{reimers2020making} to efficiently recover sentence-level representations. Specifically, given a teacher model $\mathbf{T}(\cdot)$, it maps sentences from one or more source languages $x$ into a shared dense vector space. In addition, we leverage translated sentences (which may span multiple languages) $y$ , and minimize the following objective:
\begin{equation}
    \mathcal{L}_{\text{4}} = \left[ \left( \mathbf{T}(x) - \mathbf{S}(x) \right)^2 + \left( \mathbf{T}(x) - \mathbf{S}(y) \right)^2 \right],
\end{equation}
where $\mathbf{S}(\cdot)$ denotes the student model. At this stage, the construction procedure for the multilingual parallel corpus follows the same pipeline as that used during the continued pre-training phase, resulting in a total of approximately 6.82 million multilingual sentence pairs.

\section{Experiments}
\subsection{Datasets}\label{subsec:datasets}
\textbf{Word Alignment Datasets.} We conduct experiments on human-annotated word alignment datasets covering nine language pairs: German–English (de–en), French–English (fr–en), Romanian–English (ro–en), Japanese–English (ja–en), Chinese–English (zh–en), Spanish–English (es–en), Portuguese–English (pt–en), Russian–English (ru–en), and Italian–English (it–en). The ja–en dataset is drawn from the word alignment corpus of the Kyoto Free Translation Task (KFTT)\footnote{\url{http://www.phontron.com/kftt}}, with data splits following the protocol described in \citep{wu2023wspalign}. We use all eight development files for supervised fine-tuning, while four of the seven test files are used for evaluation; the remaining three test files are reserved for future studies. The ro–en and fr–en datasets are provided by \citep{mihalcea2003evaluation}, and the de–en dataset is obtained from \citep{vilar2006aer}. For the de–en, ro–en, and fr–en language pairs, we partition the original data into training and test sets, using 300 sentence pairs for supervised fine-tuning in the fr–en and de–en datasets and 150 sentence pairs in the ro–en dataset; all remaining sentence pairs are used for evaluation. The zh–en dataset is collected from the TsinghuaAligner website\footnote{\url{http://nlp.csai.tsinghua.edu.cn/~ly/systems/TsinghuaAligner/TsinghuaAligner.html}}. We adopt the test set used in \citep{dou2021word} for evaluation, while sentence pairs outside the test set are used for supervised fine-tuning. In addition, the es–en, pt–en, ru–en, and it–en datasets are sourced from \citep{martelli2023xl}, where the development sets are used for supervised fine-tuning and the test sets are used for evaluation. Statistics of all datasets are summarized in Table~\ref{tab:Word_align_datasets}.

\begin{table}[htbp]
\centering
\caption{Statistics of the word alignment dataset, where "Avg. Tokens" denotes the statistics for the English source text in the test set.}
\label{tab:Word_align_datasets}
    \begin{tabular}{cccccccccc}
    \toprule
                & zh-en & de-en & fr-en & ro-en & ja-en & es-en & pt-en & ru-en & it-en \\ \hline
    Train Sents & 40.7K & 300   & 300   & 150   & 653   & 105   & 105   & 90    & 103   \\
    Test Sents  & 450   & 208   & 147   & 98    & 582   & 245   & 245   & 210   & 243   \\
    Avg. Tokens & 37    & 21    & 16    & 30    & 27    & 19    & 19    & 13    & 19 \\ 
    \bottomrule
    \end{tabular}

\end{table}

\textbf{Sentence Alignment Datasets.} We evaluate sentence-level alignment performance on test sets spanning seven language directions, including Chinese–English (zh–en), Spanish–English (es–en), Italian–English (it–en), German–English (de–en), French–English (fr–en), Russian–English (ru–en), and German–French (de–fr). The zh-en test set is drawn from the political (pol) domain of the aligner-eval dataset\footnote{\url{https://github.com/bfsujason/aligner-eval}}. The es–en, it–en, de–en, and fr–en test sets are obtained from \citep{molfese2024crocoalign}. In addition, the de–fr test set is sourced from the Text+Berg corpus \citep{volk2010challenges}.

\subsection{Implementation}
We initialize our model with mGTE\footnote{\url{https://huggingface.co/Alibaba-NLP/gte-multilingual-mlm-base}}, which consists of a 12-layer Transformer encoder with a hidden size of 768, and perform continued pre-training for one epoch. During this stage, the batch size is set to 512, the maximum sequence length follows the original setting of 8192, and the learning rate is set to $2e^{-5}$. A held-out validation set of $10^{4}$ samples is used for model selection. 

The self-supervised learning stage adopts the same hyperparameter configuration as the continued pre-training stage. We hold out 500 samples as the validation set and perform one epoch of fine-tuning on the model. Based on empirical results, we induce alignments using representations from the 7th Transformer layer, with the alignment threshold set to 0.001, following the method described in \citep{dou2021word}.

In the final five epochs of supervised fine-tuning for word alignment, the batch size is reduced to 64 and the learning rate is increased to $1e^{-4}$. The same alignment induction strategy is applied, using the 7th layer representations and an alignment threshold of 0.001.

To recover sentence-level representation capability, we perform Sentence Embedding Knowledge Distillation in the final stage. During this phase, parameters related to word alignment are frozen, including the embedding layer and the first seven Transformer layers, while only the last five layers are fine-tuned for five epochs. We adopt LaBSE\footnote{\url{https://huggingface.co/sentence-transformers/LaBSE}} as the teacher model. The batch size for this stage is set to 256, the validation set contains 1,000 samples, and the learning rate is again set to $2e^{-5}$.

\subsection{Baselines}
To evaluate the effectiveness of OmniAlign, we conduct experiments on both word alignment and sentence alignment tasks.

\textbf{Word Alignment Baselines.} We consider two categories of word alignment baselines. Statistical approaches include FastAlign \citep{dyer2013simple} and GIZA++ \citep{och2003systematic}, two classical and widely used models implemented under the IBM Model framework. Pre-trained language model-based approaches comprise SimAlign \citep{sabet2020simalign}, Awesome-align \citep{dou2021word}, AccAlign \citep{wang2022multilingual}, WSPAlign \citep{wu2023wspalign}, and BinaryAlign \citep{latouche2024binaryalign}. For WSPAlign (Bilingual), evaluation is conducted only on the de-en, fr-en, ro-en, ja-en, and zh-en language pairs, using checkpoints trained separately for each language direction. WSPAlign (Multilingual) is evaluated using its released multilingual checkpoint. For BinaryAlign (Bilingual), we evaluate checkpoints trained individually for each language direction.

\textbf{Sentence Alignment Baselines.} For the sentence-level alignment task, we compare OmniAlign with six existing sentence alignment tools: Gale–Church \citep{gale1993program}, VecAlign \citep{thompson2019vecalign}, BleuAlign \citep{sennrich2011iterative} (using translations generated by WPS iCiba\footnote{\url{https://www.iciba.com/translate}}), BertAlign \citep{liu2023bertalign}, SentAlign \citep{steingrimsson2023sentalign}, and CrocoAlign \citep{molfese2024crocoalign}. All methods are evaluated under their default configurations.

\subsection{Main Results}
\textbf{Word Alignment Results.} Table~\ref{tab:Word_align_results} presents a comparison between OmniAlign and various baseline methods. Using a single shared model checkpoint, OmniAlign achieves highly competitive overall performance across nine language pairs, ranking first on four datasets and second on three others. Under the setting where a single model weight is shared across all test sets, OmniAlign consistently outperforms comparable methods such as SimAlign, Awesome-align, AccAlign, and WSPAlign (Multilingual), while achieving performance comparable to the finely tuned WSPAlign (Bilingual) model. Although BinaryAlign (Bilingual) attains state-of-the-art (SOTA) results on the tasks it participates in under a fully supervised training regime, our empirical analysis indicates that its generalization capability is largely confined to the distribution of the supervised data. When evaluated on out-of-distribution samples, particularly long-form texts, its performance degrades substantially (see Table \ref{tab:word-align-long-text}). In addition, BinaryAlign requires training a separate checkpoint for each language direction, which incurs higher costs in terms of training, deployment, and maintenance in practical applications.
\begin{table}[htbp]
\small 
\setlength{\tabcolsep}{4pt} 
\centering
\caption{AER of the test set under different word alignment methods. Lower AER indicates better performance. For each language pair, the best result is \textbf{bolded} and the second-best one is \underline{underlined}.}
\label{tab:Word_align_results}
    \begin{tabular}{cccccccccc}
    \toprule
    \textbf{Methods} & \textbf{zh-en} & \textbf{de-en} & \textbf{fr-en} & \textbf{ro-en} & \textbf{ja-en} & \textbf{es-en} & \textbf{pt-en} & \textbf{ru-en} & \textbf{it-en} \\ \hline
    FastAlign \citep{dyer2013simple}               & 38.1  & 27.0  & 10.5  & 27.0    & 51.1  & -     & -     & -     & -     \\
    GIZA++ \citep{och2003systematic}                  & 35.1  & 20.6  & 5.9   & 26.4  & 48.0  & -     & -     & -     & -     \\
    SimAlign \citep{sabet2020simalign}                & 21.6  & 16.6  & 7.5   & 22.3  & 46.6  & 14.2  & 14.1  & 15.4  & 17.7  \\
    AwesomeAlign \citep{dou2021word}            & 13.3  & 13.3  & 3.8   & 18.7  & 37.4  & 12.0  & 12.7  & 13.5  & 15.7  \\
    AccAlign \citep{wang2022multilingual}                & 11.5  & 12.1  & 2.8   & 16.9  & 36.8  & \underline{11.1}  & \underline{12.1}  &                                   \underline{12.5}  & \underline{14.3}  \\
    WSPAlign (Bilingual) \citep{wu2023wspalign}    & 13.1  & 11.1  & 2.8   & \underline{10.1}  & \underline{19.3}  & -     & -     & -     & -     \\
    WSPAlign (Multilingual) \citep{wu2023wspalign} & 22.3  & 20.0  & 12.8  & 26.4  & 45.8  & 13.4  & 12.3  & 13.1  & 17.1  \\
    BinaryAlign (Bilingual) \citep{latouche2024binaryalign} & \textbf{4.8}   & \textbf{7.8}   & \textbf{1.9}   & \textbf{7.4}   & \textbf{14.3}  & -     & -     & -     & -     \\
    OmniAlign (ours)        & \underline{8.5}   & \underline{11.0}  & \underline{2.7}   & 16.7  & 29.6  & \textbf{10.7}  & \textbf{11.9}  & \textbf{12.1}                           & \textbf{14.1}  \\
    \bottomrule
    \end{tabular}
\end{table}

\textbf{Sentence Alignment Results.} Table~\ref{tab:sent_align_results} presents a comparison between OmniAlign and other sentence alignment methods. For fair comparison, all baseline methods that rely on sentence encoders employ LaBSE as their underlying representation model. Although OmniAlign acquires its sentence-level representation capability through knowledge distillation from LaBSE, the proposed model combined with our alignment algorithm achieves highly competitive overall performance, ranking first on four test sets and second on two others across seven evaluation datasets. Benefiting from the use of a token-length–based constraint rather than a character-length–based one during sentence alignment, OmniAlign achieves particularly strong performance on the zh–en test set, while exhibiting relatively weaker results on de–fr. It is worth noting that, for certain language pairs, some baseline methods still demonstrate superior performance. For example, BertAlign, leveraging its advanced alignment algorithm together with the LaBSE encoder, achieves the best scores on the en–it, en–ru, and de–fr language pairs. However, due to its modified cosine similarity function, which tends to favor many-to-many alignments, BertAlign performs comparatively worse on en–es, en–de, and en–fr.
\begin{table}[htbp]
\centering
\caption{F1 scores of OmniAlign and sentence alignment baseline methods across language pairs. For each language pair, the best result is \textbf{bolded} and the second-best one is \underline{underlined}.}
\label{tab:sent_align_results}
    \begin{tabular}{cccccccc}
    \toprule
    \textbf{Algorithm} & \textbf{en-zh} & \textbf{en-es} & \textbf{en-it} & \textbf{en-de} & \textbf{en-fr} & \textbf{en-ru} & \textbf{de-fr} \\ \hline
    Gale–Church \citep{gale1993program}        & 0.682          & \underline{0.900}    & 0.977          & 0.897          & 0.838          & 0.911          & 0.680          \\
    BleuAlign \citep{sennrich2011iterative}          & 0.782          & 0.819          & 0.901          & 0.806          & 0.757          & 0.791          & 0.770          \\
    VecAlign \citep{thompson2019vecalign}           & 0.957          & 0.892          & 0.956          & 0.869          & 0.880          & 0.921          & 0.902          \\
    BertAlign \citep{liu2023bertalign}          & \underline{0.969}    & 0.897          & \textbf{0.984} & \underline{0.900}    & \underline{0.909}    & \textbf{0.938} & \textbf{0.939} \\
    SentAlign \citep{steingrimsson2023sentalign}          & 0.968          & 0.872          & \underline{0.978}    & 0.892          & 0.903          & 0.920          & \underline{0.932}    \\
    CrocoAlign \citep{molfese2024crocoalign}         & 0.660          & 0.696          & 0.864          & 0.804          & 0.788          & 0.783          & 0.714          \\
    OmniAlign (Ours)   & \textbf{0.970} & \textbf{0.906} & \underline{0.978}    & \textbf{0.913} & \textbf{0.912} & \underline{0.935}    & 0.922          \\ \bottomrule
    \end{tabular}
\end{table}

\subsection{Ablation Study}
In this section, we compare the performance of models at different training stages, the effectiveness of embedding extraction from different layers, and the performance of various embedding models in the sentence alignment task. For the sake of brevity, we abbreviate each training stage as follows:
\begin{itemize}[left=2pt, itemsep=2pt, topsep=4pt]
    \item Continued Pre-training: S.1
    \item Self-supervised Learning: S.2
    \item Supervised Fine-tuning: S.3
    \item Sentence Embedding Knowledge Distillation: S.4
\end{itemize}

\subsubsection{Models from Different Training Stages}
Table~\ref{tab:ablation_results} reports the performance of models trained under different training strategies on the sequence alignment subtasks, evaluated across multiple test sets.

In the word alignment task, model performance improves consistently with the introduction of additional training stages. Specifically, the average alignment error rate (AER) decreases from 43.0\% to 19.6\%, 14.3\%, and finally 8.5\%, indicating that each training stage contributes positively to word-level alignment capability. Although removing S.1 and S.2 in ablation studies does not lead to a significant degradation in the final evaluation metric, both stages play an important role in shaping the model’s overall capacity. Through unsupervised or self-supervised training, the model can effectively leverage long-text samples without relying on large-scale human-annotated data, thereby improving its long-sequence modeling ability. In addition, the analysis in Table \ref{table: test_scenario} further shows that S.1 helps enhance the model’s foundational representations, enabling it to better handle word alignment tasks in complex scenarios.

For the sentence alignment task, we fix the dynamic programming (DP) algorithm and fine-tune only the last five layers of the model to align multilingual sentence representations into the English embedding space of LaBSE. Under this setting, the resulting alignment performance is comparable to that of LaBSE, and even surpasses it on the EN–ES test set. This improvement may be attributed to the multilingual joint alignment mechanism, which allows languages with relatively fewer resources to learn semantic representations more closely aligned with English. Furthermore, benefiting from the multilingual joint distillation strategy, OmniAlign outperforms several mainstream multilingual embedding models on sentence alignment tasks, despite operating with a substantially smaller parameter budget and embedding dimensionality. These results highlight the advantages of OmniAlign in terms of both parameter efficiency and computational efficiency.
\begin{table}[h]
\centering

    \begin{subtable}{\linewidth}
    \centering
    \begin{tabular}{lccc}
    \toprule
    \textbf{Model} & \textbf{zh-en} & \textbf{de-en} & \textbf{fr-en} \\
    \midrule \midrule
    \multicolumn{4}{l}{Additive Training Stages} \\
    \midrule
    mGTE-MLM-Base & 43.0 & 28.3 & 14.5 \\
    mGTE-MLM-Base + S.1 & 19.6 & 16.4 & 5.8 \\
    mGTE-MLM-Base + S.1 + S.2 & \underline{14.3} & \underline{12.8} & \underline{3.4} \\
    mGTE-MLM-Base + S.1 + S.2 + S.3 & \textbf{8.5} & \textbf{11.0} & \textbf{2.7} \\
    \midrule \midrule
    \multicolumn{4}{l}{Subtractive Training Stages} \\
    \midrule
    OmniAlign w/o S.3 & 14.3 & 12.8 & 3.4 \\
    OmniAlign w/o S.2 & \underline{8.6} & \underline{11.1} & \textbf{2.7} \\
    OmniAlign w/o S.1 & 8.7 & 11.5 & \underline{2.9} \\
    OmniAlign & \textbf{8.5} & \textbf{11.0} & \textbf{2.7} \\
    \bottomrule
    \end{tabular}
    \caption{Word alignment AER (\%). Lower is better. Within each block, the best result is \textbf{bolded} and the second-best one is \underline{underlined}.}
    \end{subtable}

\vspace{0.6em}

    \begin{subtable}{\linewidth}
    \centering
    \begin{tabular}{lccc ccc}
    \toprule
    \textbf{Method} & \textbf{\#Params} & \textbf{Emb. Dim.} & \textbf{en-zh} & \textbf{en-es} & \textbf{en-it} \\
    \midrule
    DP + LaBSE & 0.47B & 768 & \textbf{0.972} & 0.904 & \textbf{0.979} \\
    DP + multilingual-e5-large-instructE\footnote{\url{https://huggingface.co/intfloat/multilingual-e5-large-instruct}} & 0.56B & 1024 & 0.895 & 0.899 & 0.972 \\
    DP + bge-m3\footnote{\url{https://huggingface.co/BAAI/bge-m3}} & 0.57B & 1024 & 0.933 & \textbf{0.910} & \underline{0.978} \\
    DP + Qwen3-Embedding-0.6B\footnote{\url{https://huggingface.co/Qwen/Qwen3-Embedding-0.6B}} & 0.6B & 1024 & 0.939 & 0.905 & 0.971 \\
    DP + Qwen3-Embedding-8B\footnote{\url{https://huggingface.co/Qwen/Qwen3-Embedding-8B}} & 8B & 4096 & 0.935 & \underline{0.908} & 0.975 \\
    DP + gte-multilingual-base\footnote{\url{https://huggingface.co/Alibaba-NLP/gte-multilingual-base}} & 0.31B & 768 & 0.927 & \underline{0.908} & 0.973 \\
    DP + OmniAlign (w/o S.4) & 0.31B & 768 & 0.716 & 0.872 & 0.966 \\
    DP + OmniAlign (distilled from LaBSE) & 0.31B & 768 & \underline{0.970} & 0.906 & \underline{0.978} \\
    \bottomrule
    \end{tabular}
    \caption{Sentence alignment F1 scores. Higher is better. For each language pair, the best result is \textbf{bolded} and the second-best one is \underline{underlined}.}
    \end{subtable}

\caption{Ablation study on word and sentence alignment performance under different training strategies and embedding models.}
\label{tab:ablation_results}
\end{table}

\subsubsection{Word Alignment Performance across Different Word Embedding Layers}\label{subsec:3.5.2}
In this section, we investigate the word alignment performance of mBERT, mGTE-MLM-Base, and its continued pre-trained variant across different Transformer layers, all of which are widely used in word alignment tasks. Experimental results show that, as a base model, mGTE-MLM-Base underperforms mBERT in terms of word alignment accuracy. However, mGTE-MLM-Base offers a key advantage in that its native configuration supports a substantially longer maximum sequence length than mBERT (8192 vs. 512), and this capacity can be further extended through RoPE-based scaling techniques, such as positional frequency scaling or linear interpolation \citep{chen2023extending, peng2023yarn}.

Motivated by this property, we perform task-oriented continued pre-training on mGTE-MLM-Base specifically for word alignment, aiming to compensate for its initial shortcomings in word-level alignment performance while preserving its long-context modeling capability.
\begin{table}[htbp]
\centering
\small
\begin{tabular}{llccc}
\toprule
\textbf{Model} & \textbf{Layer} & \textbf{zh-en} & \textbf{de-en} & \textbf{fr-en} \\
\midrule
\multirow{3}{*}{mBERT}
 & 6 & 20.5 & 19.4 & 6.5 \\
 & 7 & 19.0 & 16.6 & 5.6 \\
 & 8 & \textbf{18.1} & \textbf{15.2} & \textbf{5.4} \\
\midrule
\multirow{3}{*}{mGTE-MLM-Base}
 & 6 & 54.8 & 33.7 & 18.0 \\
 & 7 & \textbf{43.0} & \textbf{28.3} & \textbf{14.5} \\
 & 8 & 47.5 & 30.1 & 16.0 \\
\midrule
\multirow{3}{*}{mGTE-MLM-Base + S.1}
 & 6 & 28.4 & 21.2 & 6.9 \\
 & 7 & \textbf{19.6} & \textbf{16.4} & \textbf{5.8} \\
 & 8 & 22.8 & 19.1 & 6.4 \\
\bottomrule
\end{tabular}
\caption{Word alignment AER (\%) at different layers for various base models. Within each model group, the best result is \textbf{bolded}.}
\label{tab:layerwise_alignment}
\end{table}

\subsubsection{Exploring Long Text Word Alignment Capability}
As shown in Table \ref{tab:Word_align_datasets}, existing evaluations for word alignment tasks are predominantly conducted in short-text settings. To systematically assess model performance under long-text word alignment conditions, we extend the original test sets to long-context scenarios using a set of heuristic rules. Specifically, zh–en–n denotes the construction of longer input sequences by concatenating n samples from the original zh–en test set.

The experimental results in Table \ref{tab:word-align-long-text} show that, as the input text length increases, the performance of previous word alignment methods degrades substantially. Moreover, these methods are typically constrained by a maximum input length of 512 tokens, which prevents them from effectively supporting word alignment in longer sequence settings. In contrast, OmniAlign degrades only mildly, with AER rising from 8.5 on zh–en–1 to 12.6 on zh–en–50. This robustness comes from a long-context backbone that can encode up to 8,192 tokens in one pass, and from early-stage training on mixed-length—especially long—texts. Supervised fine-tuning on short annotated data further improves alignment quality, and these gains also hold under long-text evaluation.
\begin{table}[htbp]
\centering
\begin{tabular}{lcccccc}
\toprule
\textbf{Model} & \textbf{zh-en-1} & \textbf{zh-en-3} & \textbf{zh-en-5} & \textbf{zh-en-10} & \textbf{zh-en-20} & \textbf{zh-en-50} \\
\hline
Awesome-Align          & 13.3 & 13.9 & 15.1 & 24.8 & --   & --   \\
ACC-Align              & 11.5 & 12.5 & 15.8 & 28.1 & --   & --   \\
BinaryAlign (Bilingual)&  4.8 &  9.6 & 22.8 & --   & --   & --   \\
OmniAlign (Ours)       &  8.5 &  8.6 &  8.6 &  9.2 & 10.5 & 12.6 \\
\bottomrule
\end{tabular}
\caption{Word alignment performance under increasing input lengths. The zh-en-$n$ setting denotes long-text inputs constructed by concatenating $n$ samples from the original zh-en test set. Notably, zh–en–50 reaches 1,850 tokens.
}
\label{tab:word-align-long-text}
\end{table}

\subsubsection{Sequence alignment on unseen language pairs}
To evaluate the zero-shot generalization capability of OmniAlign, we directly apply all methods to previously unseen language pairs, without performing any additional training on the target test languages. For the word alignment task, we conduct experiments on the bg–en, da–en, et–en, hu–en, nl–en, and sl–en language pairs, using test datasets sourced from \citep{martelli2023xl}. For the sentence alignment task, we evaluate on the en–hu and en–nl test sets, which are obtained from \cite{molfese2024crocoalign}.

Table \ref{tab:unseen-language-pairs-align} reports the alignment performance of different methods on these unseen language pairs. OmniAlign matches or outperforms the baselines on most pairs. We attribute this to training a single shared multilingual encoder rather than language-pair-specific aligners: continued pre-training and self-supervised learning on large-scale multilingual parallel data induce token-level correspondences that are not tied to a particular language pair, while supervised fine-tuning on seen pairs further sharpens these signals and transfers them to unseen directions.
\begin{table}[htbp]
\centering
\begin{minipage}{0.62\linewidth}
\centering
\subcaption{Word alignment AER}
\begin{tabular}{lccc}
\toprule
Test Set & AccAlign & BinaryAlign & OmniAlign \\
\hline
bg-en & 11.0 & 11.99 & 10.6 \\
da-en & 7.1 & 8.45 & 7.1 \\
et-en & 15.2 & 15.67 & 14.4 \\
hu-en & 18.7 & 16.93 & 19.1 \\
nl-en & 4.6 & 5.14 & 4.5 \\
sl-en & 14.9 & 16.85 & 14.7 \\
\bottomrule
\end{tabular}
\end{minipage}
\hfill
\begin{minipage}{0.34\linewidth}
\centering
\subcaption{Sentence alignment F1 score}
\begin{tabular}{lcc}
\toprule
Test Set & BertAlign & Ours \\
\hline
en-hu & 0.977 & 0.979 \\
en-nl & 0.928 & 0.933 \\
\bottomrule
\end{tabular}
\end{minipage}
\caption{Alignment performance of OmniAlign and baselines on unseen language pairs}
\label{tab:unseen-language-pairs-align}
\end{table}

\subsection{Case Study}
To complement the quantitative benchmarks, we present qualitative case studies in Table~\ref{table: test_scenario}, covering both word-level and sentence-level alignment under challenging conditions. 
\textbf{Scenario 1} focuses on word alignment. In Example \#1, the key difficulty is mapping an English noun such as ``failure'' to a Chinese negated expression (``没有成功''). Models without continued pre-training (OmniAlign w/o S.1) often align only the positive lexical item and miss the negation scope. Example \#2 examines function-word and numeral expressions in a medical instruction sentence (e.g., ``有'' and ``二岁''), where partial matches can look locally plausible but remain incomplete. Example \#3 involves informal user-generated text with morphological variation and mild noise (e.g., ``returned'' and the misspelling ``windering''), which stresses robustness beyond clean parallel news text. Example \#4 concatenates a long technical paragraph with many multi-token named entities and domain phrases; under this setting, BinaryAlign fails to produce usable alignment outputs, whereas OmniAlign still recovers the highlighted correspondences.

\textbf{Scenario 2} turns to document-level sentence alignment, where the central issue is choosing an appropriate $m$--$n$ segmentation rather than maximizing span coverage. Across Examples \#1--\#4, BertAlign frequently merges neighboring sentences into broader many-to-many blocks, even when finer one-to-one or constrained $m$--$n$ links are preferable. OmniAlign, guided by anchor-constrained dynamic programming with a token-length--based penalty, tends to produce more compact and interpretable alignment units on these cases. 
\begin{center}
	\scriptsize
	\renewcommand{\arraystretch}{1.4}
	\setlength{\tabcolsep}{3pt}
	\begin{longtable}{p{3cm}p{12.5cm}}  
		\caption{Case studies of OmniAlign and other sequence alignment methods. \textcolor{blue}{Blue} indicates correct alignments, \textcolor{red}{red} denotes incorrect alignments, and black marks segments with no alignment.} \\
		\hline
		\multicolumn{2}{c}{\textbf{Scenario 1: Word Alignment}} \\
		\hline
		
		\textbf{Example \#1} & He had to repeatedly traverse the inner palace's chambers, yet each attempt ended in \textcolor{blue}{\textbf{failure}}. \\
		\textbf{Testing Points} & “failure”. \\
		\textbf{OmniAlign} & 他得不断地一再地穿过内宫里的屋子；可是他一直\textcolor{blue}{\textbf{没有成功}}。\\
		\textbf{OmniAlign w/o S.1} & 他得不断地一再地穿过内宫里的屋子；可是他一直没有\textcolor{red}{\textbf{成功}}。 \\
		\hline
		
		\textbf{Example \#2} & 已知\textcolor{blue}{\textbf{有}}血液疾病及尿酸性肾结石的患者不推荐使用本品，\textcolor{blue}{\textbf{二岁}}以下儿童不得服用。 \\
		\textbf{Testing Points} & 有; 二岁 \\
		\textbf{OmniAlign} & This product is not recommended for patients \textcolor{blue}{\textbf{with}} known blood disorders or uric acid kidney stones, and it should not be taken by children under the \textcolor{blue}{\textbf{age}} of \textcolor{blue}{\textbf{two}}. \\
		\textbf{BinaryAlign} & This product is not recommended for patients with known blood disorders or uric acid kidney stones, and it should not be taken by children under the \textcolor{red}{\textbf{age}} of two. \\
		\hline

		\textbf{Example \#3} & I recently \textcolor{blue}{\textbf{returned}} to d2 after several year, now I’m  \textcolor{blue}{\textbf{windering}}: Where do you guys sell/buy your stuff? Do you just make a game “O xxx N yyy” and hope for the best? Or are there a website that’s more efficient? \\
		\textbf{Testing Points} & returned; windering \\
		\textbf{OmniAlign} & 时隔多年，我又 \textcolor{blue}{\textbf{重新}}开始玩《暗黑破坏神2》（DiabloII），现在我 \textcolor{blue}{\textbf{想知道}}：大家都是在哪里进行物品交易的？是通过自己创建名为“OxxxNyyy”的游戏房间来交易，然后听天由命吗？还是有更高效的交易网站？ \\
		\textbf{BinaryAlign} & 时隔多年，我\textcolor{red}{\textbf{又重新开始玩}}《暗黑破坏神2》（DiabloII），现在我想知道：大家都是在哪里进行物品交易的？是通过自己创建名为“OxxxNyyy”的游戏房间来交易，然后\textcolor{red}{\textbf{听天由命}}吗？还是有更高效的交易网站？ \\
		\hline        

		\textbf{Example \#4} & 在技术演进的脉络中，预训练模型的出现无疑是里程碑式的突破。以 \textcolor{blue}{\textbf{BERT、GPT、T5}} 等为代表的模型通过在海量无标注文本上进行自监督学习，习得语言的深层语义表示和语法结构，极大地提升了下游任务的性能上限。这些模型基于 Transformer 架构的\textcolor{blue}{\textbf{多头注意力机制}}，能够捕捉文本中的长距离依赖关系和上下文关联，为\textcolor{blue}{\textbf{文本分类、情感分析、机器翻译、问答系统}}等经典任务提供了强大的通用能力。近年来，多语言预训练模型（如 \textcolor{blue}{\textbf{XLM-R、mT5、LaBSE}}）进一步打破了语言壁垒，通过对数百种语言的联合训练，实现了跨语言文本的理解与生成，为全球化信息传播和跨文化交流提供了技术支撑。在产业应用层面，NLP 技术已渗透到金融、医疗、教育、传媒等多个领域，催生了一系列\textcolor{blue}{\textbf{创新产品和服务}}。在金融行业，\textcolor{blue}{\textbf{智能客服系统}}能够基于用户的自然语言查询快速提供账户咨询、业务办理等服务，平均响应时间缩短至秒级，客户满意度提升 30\% 以上；\textcolor{blue}{\textbf{风险控制模型}}通过分析企业年报、新闻舆情等文本数据，精准识别信用风险和市场波动信号，帮助金融机构降低不良贷款率。在医疗领域，医学文本分析系统可自动提取电子病历中的关键信息（如病症、用药史、检查结果），辅助医生进行诊断决策，减少误诊率；\textcolor{blue}{\textbf{多语言医疗翻译工具}}则为跨境医疗合作提供了语言保障，使不同国家的医护人员能够高效协作。在教育领域，智能写作辅助系统能够\textcolor{blue}{\textbf{实时检测}}文本中的语法错误、逻辑漏洞，并提供优化建议，帮助学生提升写作能力；个性化学习平台通过分析学生的\textcolor{blue}{\textbf{学习行为和文本交互数据}}，精准推送适配的学习资源，实现 “因材施教” 的 \textcolor{blue}{\textbf{教育理念}}。 \\
		\textbf{Testing Points} & (Bert、GPT、T5); (多头注意力机制); (文本分类、情感分析、机器翻译、问答系统); (XLM-R、mT5、LaBSE); (创新产品和服务); (智能客服系统); (风险控制模型); (多语言医疗翻译工具); (实时检测); (学习行为和文本交互数据); (教育理念) \\
		\textbf{OmniAlign} & In the context of technological evolution, the emergence of pre-trained models is undoubtedly a milestone breakthrough. Represented by \textcolor{blue}{\textbf{BERT, GPT, T5}}, and other models, they acquire deep semantic representations and syntactic structures of language through self-supervised learning on massive unlabeled text, greatly raising the performance ceiling of downstream tasks. Based on the \textcolor{blue}{\textbf{multi-head attention mechanism}} of the Transformer architecture, these models can capture long-distance dependencies and contextual correlations in text, providing powerful general capabilities for classic tasks such as \textcolor{blue}{\textbf{text classification, sentiment analysis, machine translation, and question-answering systems}}. In recent years, multilingual pre-trained models (e.g., \textcolor{blue}{\textbf{XLM-R, mT5, LaBSE}}) have further broken down language barriers. Through joint training on hundreds of languages, they have realized the understanding and generation of cross-lingual text, providing technical support for global information dissemination and cross-cultural communication. At the industrial application level, NLP technology has penetrated into multiple fields such as finance, healthcare, education, and media, spawning a series of \textcolor{blue}{\textbf{innovative products and services}}. In the financial industry, \textcolor{blue}{\textbf{intelligent customer service systems}} can quickly provide account consultations, business processing, and other services based on users' natural language queries, reducing the average response time to seconds and increasing customer satisfaction by more than 30\%; \textcolor{blue}{\textbf{risk control models}} accurately identify credit risks and market fluctuation signals by analyzing text data such as corporate annual reports and news public opinion, helping financial institutions reduce the non-performing loan ratio. In the healthcare field, medical text analysis systems can automatically extract key information from electronic medical records (such as symptoms, medication history, and examination results) to assist doctors in diagnostic decisions and reduce the misdiagnosis rate; \textcolor{blue}{\textbf{multilingual medical translation tools}} provide language guarantees for cross-border medical cooperation, enabling medical staff from different countries to collaborate efficiently. In the education field, intelligent writing assistance systems can \textcolor{blue}{\textbf{real-time detect}} grammatical errors and logical flaws in text, and provide optimization suggestions to help students improve their writing skills; personalized learning platforms accurately push adaptive learning resources by analyzing students' \textcolor{blue}{\textbf{learning behaviors and text interaction data}}, realizing the \textcolor{blue}{\textbf{educational concept}} of "teaching students in accordance with their aptitude". \\
		\textbf{BinaryAlign} & In the context of technological evolution, the emergence of pre-trained models is undoubtedly a milestone breakthrough. Represented by BERT, GPT, T5, and other models, they acquire deep semantic representations and syntactic structures of language through self-supervised learning on massive unlabeled text, greatly raising the performance ceiling of downstream tasks. Based on the multi-head attention mechanism of the Transformer architecture, these models can capture long-distance dependencies and contextual correlations in text, providing powerful general capabilities for classic tasks such as text classification, sentiment analysis, machine translation, and question-answering systems. In recent years, multilingual pre-trained models (e.g., XLM-R, mT5, LaBSE) have further broken down language barriers. Through joint training on hundreds of languages, they have realized the understanding and generation of cross-lingual text, providing technical support for global information dissemination and cross-cultural communication. At the industrial application level, NLP technology has penetrated into multiple fields such as finance, healthcare, education, and media, spawning a series of innovative products and services. In the financial industry, intelligent customer service systems can quickly provide account consultations, business processing, and other services based on users' natural language queries, reducing the average response time to seconds and increasing customer satisfaction by more than 30\%; risk control models accurately identify credit risks and market fluctuation signals by analyzing text data such as corporate annual reports and news public opinion, helping financial institutions reduce the non-performing loan ratio. In the healthcare field, medical text analysis systems can automatically extract key information from electronic medical records (such as symptoms, medication history, and examination results) to assist doctors in diagnostic decisions and reduce the misdiagnosis rate; multilingual medical translation tools provide language guarantees for cross-border medical cooperation, enabling medical staff from different countries to collaborate efficiently. In the education field, intelligent writing assistance systems can real-time detect grammatical errors and logical flaws in text, and provide optimization suggestions to help students improve their writing skills; personalized learning platforms accurately push adaptive learning resources by analyzing students' learning behaviors and text interaction data, realizing the educational concept of "teaching students in accordance with their aptitude". \\
		\hline
		
		\multicolumn{2}{c}{\textbf{Scenario 2: Sentence Alignment}} \\
		\hline
		
        \textbf{Example \#1} &
        \begin{minipage}[t]{\linewidth}
        \textbf{Source} \\
        \hangindent=1.2em
        (1) Their Australian born captain, the world's top-ranked match racer Peter Gilmour, lived in Japan for three years to satisfy the Cup's crew-nationality rules. \\
        \hangindent=1.2em
        (2) Syndicate head Tatsumitsu Yamasaki, chairman of spice giant S\&B Foods, is hungry for a win. \\[0.4ex]
        
        \textbf{Target} \\
        \hangindent=1.2em
        (1) 该船队的船长为世界顶尖帆船赛选手皮得·吉尔摩。 \\
        \hangindent=1.2em
        (2) 为了达到杯赛在船员国籍方面的各项要求，这位出生在澳大利亚的选手在日本居住了3年。 \\
        \hangindent=1.2em
        (3) 山崎达光是一家财团的总裁兼调味品行业的巨人S\&B食品株式会社董事会主席。 \\
        \hangindent=1.2em
        (4) 他渴望日本船队能够获得胜利。
        \end{minipage}
        \\

		\textbf{OmniAlign} & \textcolor{blue}{[(1)]:[(1), (2)]; [(2)]:[(3), (4)]}\\
		\textbf{BertAlign} & \textcolor{blue}{[(1)]:[(1), (2)]}; \textcolor{red}{[(2)]:[(3)]; []:[(4)]}\\
		\hline
		
		\textbf{Example \#2} &
        \begin{minipage}[t]{\linewidth}
        \textbf{Source} \\
        \hangindent=1.2em
        (1) Then came the hard part: getting people to want to see the thing. \\
        \hangindent=1.2em
        (2) Here's how they did it, from whisper to buzz to big box-office noise, in only 21 steps. \\[0.4ex]
        
        \textbf{Target} \\
        \hangindent=1.2em
        (1) 随后进入最艰苦的宣传炒作阶段。 \\
        \hangindent=1.2em
        (2) 下面就是他们的宣传炒作步骤：刚开始知者甚少，随后观众渐增，最后票房炙手可热。 \\
        \hangindent=1.2em
        (3) 这一成绩的取得仅需21步：。 \\
        \end{minipage}
        \\

		\textbf{OmniAlign} & \textcolor{blue}{[(1)]:[(1), (2)]; [(2)]:[(2), (3)]}\\
		\textbf{BertAlign} & \textcolor{red}{[(1), (2)]:[(1), (2), (3)]}\\
		\hline

        \textbf{Example \#3} &
        \begin{minipage}[t]{\linewidth}
        \textbf{Source} \\
        \hangindent=1.2em
        (1) The situation in Japan has to change. " \\
        \hangindent=1.2em
        (2) At Nissan, it already has. \\[0.4ex]
        
        \textbf{Target} \\
        \hangindent=1.2em
        (1) 这种形势必须改变。\\
        \hangindent=1.2em
        (2) 在日产，这种变革已经开始。  \\
        \end{minipage}
        \\
        \textbf{OmniAlign} & \textcolor{blue}{[(1)]:[(1)]; [(2)]:[(2)]}\\
		\textbf{BertAlign} & \textcolor{red}{[(1), (2)]:[(1), (2)]}\\
		\hline

        \textbf{Example \#4} &
        \begin{minipage}[t]{\linewidth}
        \textbf{Source} \\
        \hangindent=1.2em
        (1) 就在这种情况下，我想起十五队的队医陈清扬是北医大毕业的大夫，对针头和勾针大概还能分清，所以我去找她看病。 \\
        \hangindent=1.2em
        (2) 看完病回来，不到半个小时，她就追到我屋里来，要我证明她不是破鞋。 \\[0.4ex]
        
        \textbf{Target} \\
        \hangindent=1.2em
        (1) Under the circumstances, I recalled that the doctor at the fifteenth team, Chen Qingyang, had graduated from Beijing Medical School.\\
        \hangindent=1.2em
        (2) Maybe she would be able to tell the difference between a hypodermic and a crotchet needle.  \\
        \hangindent=1.2em
        (3) So I went to see her.  \\
        \hangindent=1.2em
        (4) Not half an hour after my visit, she chased after me to my room, wanting me to prove that she wasn't damaged goods.   \\
        \end{minipage}
        \\
        \textbf{OmniAlign} & \textcolor{blue}{[(1)]:[(1), (2), (3)]; [(2)]:[(4)]}\\
		\textbf{BertAlign} & \textcolor{red}{[(1)]:[(1), (2)]; [(2)]:[(3), (4)]}\\
		\hline
		
		\label{table: test_scenario}
	\end{longtable}
\end{center}

\section{Related work}
Cross-lingual sequence alignment provides essential signals for parallel corpus construction and downstream multilingual applications \citep{varga2008parallel, david2001inducing, hashimoto2019high, tu2016modeling, nicolai2019learning, mayhew2017cheap, chi2021improving}, but prior work is often specialized to a single granularity. For word alignment, classical statistical models such as GIZA++ \citep{och2003systematic} and FastAlign \citep{dyer2013simple} remain strong baselines. Recent approaches leverage multilingual contextual representations to induce alignments, including SimAlign \citep{sabet2020simalign}, Awesome-align \citep{dou2021word}, AccAlign \citep{wang2022multilingual}, WSPAlign \citep{wu2023wspalign}, and BinaryAlign \citep{latouche2024binaryalign}. While these methods improve accuracy by exploiting pre-trained encoders and supervised signals, many are evaluated under short-context settings and can be limited by encoder input length.

For sentence/document alignment, early length-based algorithms such as Gale--Church \citep{gale1993program} align using dynamic programming, while translation-assisted methods like BleuAlign \citep{sennrich2011iterative} rely on external MT systems. Embedding-based approaches (e.g., VecAlign \citep{thompson2019vecalign}, BertAlign \citep{liu2023bertalign}, SentAlign \citep{steingrimsson2023sentalign}, CrocoAlign \citep{molfese2024crocoalign}) compute cross-lingual sentence representations—often using models such as LaBSE \citep{feng2022language} or LASER \citep{artetxe2019massively}—and combine similarity retrieval with sequence constraints. However, these methods typically focus on sentence-level alignment and do not provide fine-grained word alignments. Overall, existing tools seldom offer a unified solution that simultaneously supports multilingual, long-context, and multi-granularity alignment.

\section{Conclusion}
In this report, we present OmniAlign, a unified multilingual aligner that supports both word-level and sentence-level alignment within a single lightweight model. OmniAlign is built upon an encoder-only architecture with strong long-context modeling capability, where token- and word-level alignments are induced from contextualized representations. For document-level sentence alignment, OmniAlign further integrates a two-stage, anchor-constrained dynamic programming algorithm.

To balance fine-grained alignment accuracy and high-quality sentence representations, we design a four-stage training pipeline consisting of continued pre-training, self-supervised learning, supervised fine-tuning with human-annotated data, and sentence embedding knowledge distillation from a strong multilingual teacher model. Extensive experiments demonstrate that OmniAlign achieves highly competitive performance at both the word and sentence alignment levels. Notably, in long-text word alignment scenarios, OmniAlign maintains stable performance, whereas many existing methods suffer from severe degradation due to context-length limitations. In addition, OmniAlign exhibits strong zero-shot generalization on unseen language pairs, delivering reliable alignment results without any additional training.

\section{Acknowledgment}
We thank the Engineering and Infrastructure teams at WPS-Qingqiu, as well as the WPS-iCIBA team, for their support. In particular, we appreciate their assistance with data collection, pre-processing, and platform support.

\end{CJK*}
\bibliographystyle{colm2024_conference}
\bibliography{colm2024_conference}

@incollection{varga2008parallel,
  title={Parallel corpora for medium density languages},
  author={Varga, D{\'a}niel and Hal{\'a}csy, P{\'e}ter and Kornai, Andr{\'a}s and Nagy, Viktor and N{\'e}meth, L{\'a}szl{\'o} and Tr{\'o}n, Viktor},
  booktitle={Recent advances in natural language processing IV: selected papers from RANLP 2005},
  pages={247--258},
  year={2008},
  publisher={John Benjamins Publishing Company}
}

@inproceedings{david2001inducing,
  title={Inducing multilingual text analysis tools via robust projection across aligned corpora},
  author={David, Yarowsky and Grace, Ngai and Richard, Wicentowski and others},
  booktitle={Proceedings of the First International Conference on Human Language Technology Research},
  pages={1--8},
  year={2001}
}

@inproceedings{hashimoto2019high,
  title={A high-quality multilingual dataset for structured documentation translation},
  author={Hashimoto, Kazuma and Buschiazzo, Raffaella and Bradbury, James and Marshall, Teresa and Socher, Richard and Xiong, Caiming},
  booktitle={Proceedings of the Fourth Conference on Machine Translation (Volume 1: Research Papers)},
  pages={116--127},
  year={2019}
}

@article{tu2016modeling,
  title={Modeling coverage for neural machine translation},
  author={Tu, Zhaopeng and Lu, Zhengdong and Liu, Yang and Liu, Xiaohua and Li, Hang},
  journal={arXiv preprint arXiv:1601.04811},
  year={2016}
}

@inproceedings{nicolai2019learning,
  title={Learning morphosyntactic analyzers from the Bible via iterative annotation projection across 26 languages},
  author={Nicolai, Garrett and Yarowsky, David},
  booktitle={Proceedings of the 57th Annual Meeting of the Association for Computational Linguistics},
  pages={1765--1774},
  year={2019}
}

@inproceedings{mayhew2017cheap,
  title={Cheap translation for cross-lingual named entity recognition},
  author={Mayhew, Stephen and Tsai, Chen-Tse and Roth, Dan},
  booktitle={Proceedings of the 2017 conference on empirical methods in natural language processing},
  pages={2536--2545},
  year={2017}
}

@inproceedings{chi2021improving,
  title={Improving pretrained cross-lingual language models via self-labeled word alignment},
  author={Chi, Zewen and Dong, Li and Zheng, Bo and Huang, Shaohan and Mao, Xian-Ling and Huang, He-Yan and Wei, Furu},
  booktitle={Proceedings of the 59th Annual Meeting of the Association for Computational Linguistics and the 11th International Joint Conference on Natural Language Processing (Volume 1: Long Papers)},
  pages={3418--3430},
  year={2021}
}

@article{sabet2020simalign,
  title={SimAlign: High quality word alignments without parallel training data using static and contextualized embeddings},
  author={Sabet, Masoud Jalili and Dufter, Philipp and Yvon, Fran{\c{c}}ois and Sch{\"u}tze, Hinrich},
  journal={arXiv preprint arXiv:2004.08728},
  year={2020}
}

@article{dou2021word,
  title={Word alignment by fine-tuning embeddings on parallel corpora},
  author={Dou, Zi-Yi and Neubig, Graham},
  journal={arXiv preprint arXiv:2101.08231},
  year={2021}
}

@inproceedings{wang2022multilingual,
  title={Multilingual sentence transformer as a multilingual word aligner},
  author={Wang, Weikang and Chen, Guanhua and Wang, Hanqing and Han, Yue and Chen, Yun},
  booktitle={Findings of the Association for Computational Linguistics: EMNLP 2022},
  pages={2952--2963},
  year={2022}
}

@article{wu2023wspalign,
  title={WSPAlign: Word alignment pre-training via large-scale weakly supervised span prediction},
  author={Wu, Qiyu and Nagata, Masaaki and Tsuruoka, Yoshimasa},
  journal={arXiv preprint arXiv:2306.05644},
  year={2023}
}

@inproceedings{nagata2020supervised,
  title={A supervised word alignment method based on cross-language span prediction using multilingual BERT},
  author={Nagata, Masaaki and Chousa, Katsuki and Nishino, Masaaki},
  booktitle={Proceedings of the 2020 Conference on Empirical Methods in Natural Language Processing (EMNLP)},
  pages={555--565},
  year={2020}
}

@article{lai2022cross,
  title={Cross-align: Modeling deep cross-lingual interactions for word alignment},
  author={Lai, Siyu and Yang, Zhen and Meng, Fandong and Chen, Yufeng and Xu, Jinan and Zhou, Jie},
  journal={arXiv preprint arXiv:2210.04141},
  year={2022}
}

@inproceedings{latouche2024binaryalign,
  title={BinaryAlign: Word Alignment as Binary Sequence Labeling},
  author={Latouche, Gaetan and Carbonneau, Marc-Andr{\'e} and Swanson, Ben},
  booktitle={Proceedings of the 62nd Annual Meeting of the Association for Computational Linguistics (Volume 1: Long Papers)},
  pages={10277--10288},
  year={2024}
}

@inproceedings{molfese2024crocoalign,
  title={Crocoalign: A cross-lingual, context-aware and fully-neural sentence alignment system for long texts},
  author={Molfese, Francesco Maria and Bejgu, Andrei Stefan and Tedeschi, Simone and Conia, Simone and Navigli, Roberto},
  booktitle={Proceedings of the 18th Conference of the European Chapter of the Association for Computational Linguistics (Volume 1: Long Papers)},
  pages={2209--2220},
  year={2024}
}

@inproceedings{sennrich2011iterative,
  title={Iterative, MT-based sentence alignment of parallel texts},
  author={Sennrich, Rico and Volk, Martin},
  booktitle={Proceedings of the 18th Nordic conference of computational linguistics (NODALIDA 2011)},
  pages={175--182},
  year={2011}
}

@article{steingrimsson2023sentalign,
  title={SentAlign: Accurate and scalable sentence alignment},
  author={Steingr{\'\i}msson, Stein{\th}{\'o}r and Loftsson, Hrafn and Way, Andy},
  journal={arXiv preprint arXiv:2311.08982},
  year={2023}
}

@inproceedings{thompson2019vecalign,
  title={Vecalign: Improved sentence alignment in linear time and space},
  author={Thompson, Brian and Koehn, Philipp},
  booktitle={Proceedings of the 2019 conference on empirical methods in natural language processing and the 9th international joint conference on natural language processing (EMNLP-IJCNLP)},
  pages={1342--1348},
  year={2019}
}

@article{liu2023bertalign,
  title={Bertalign: Improved word embedding-based sentence alignment for Chinese--English parallel corpora of literary texts},
  author={Liu, Lei and Zhu, Min},
  journal={Digital Scholarship in the Humanities},
  volume={38},
  number={2},
  pages={621--634},
  year={2023},
  publisher={Oxford University Press}
}

@inproceedings{feng2022language,
  title={Language-agnostic BERT sentence embedding},
  author={Feng, Fangxiaoyu and Yang, Yinfei and Cer, Daniel and Arivazhagan, Naveen and Wang, Wei},
  booktitle={Proceedings of the 60th Annual Meeting of the Association for Computational Linguistics (Volume 1: Long Papers)},
  pages={878--891},
  year={2022}
}

@article{artetxe2019massively,
  title={Massively multilingual sentence embeddings for zero-shot cross-lingual transfer and beyond},
  author={Artetxe, Mikel and Schwenk, Holger},
  journal={Transactions of the association for computational linguistics},
  volume={7},
  pages={597--610},
  year={2019},
  publisher={MIT Press One Rogers Street, Cambridge, MA 02142-1209, USA journals-info~…}
}

@article{zhang2024mgte,
  title={mgte: Generalized long-context text representation and reranking models for multilingual text retrieval},
  author={Zhang, Xin and Zhang, Yanzhao and Long, Dingkun and Xie, Wen and Dai, Ziqi and Tang, Jialong and Lin, Huan and Yang, Baosong and Xie, Pengjun and Huang, Fei and others},
  journal={arXiv preprint arXiv:2407.19669},
  year={2024}
}

@article{pires2019multilingual,
  title={How multilingual is multilingual BERT?},
  author={Pires, Telmo and Schlinger, Eva and Garrette, Dan},
  journal={arXiv preprint arXiv:1906.01502},
  year={2019}
}

@inproceedings{conneau2020unsupervised,
  title={Unsupervised cross-lingual representation learning at scale},
  author={Conneau, Alexis and Khandelwal, Kartikay and Goyal, Naman and Chaudhary, Vishrav and Wenzek, Guillaume and Guzm{\'a}n, Francisco and Grave, Edouard and Ott, Myle and Zettlemoyer, Luke and Stoyanov, Veselin},
  booktitle={Proceedings of the 58th annual meeting of the association for computational linguistics},
  pages={8440--8451},
  year={2020}
}

@article{gururangan2020don,
  title={Don't stop pretraining: Adapt language models to domains and tasks},
  author={Gururangan, Suchin and Marasovi{\'c}, Ana and Swayamdipta, Swabha and Lo, Kyle and Beltagy, Iz and Downey, Doug and Smith, Noah A},
  journal={arXiv preprint arXiv:2004.10964},
  year={2020}
}

@article{conneau2019cross,
  title={Cross-lingual language model pretraining},
  author={Conneau, Alexis and Lample, Guillaume},
  journal={Advances in neural information processing systems},
  volume={32},
  year={2019}
}

@article{reimers2020making,
  title={Making monolingual sentence embeddings multilingual using knowledge distillation},
  author={Reimers, Nils and Gurevych, Iryna},
  journal={arXiv preprint arXiv:2004.09813},
  year={2020}
}

@inproceedings{mihalcea2003evaluation,
  title={An evaluation exercise for word alignment},
  author={Mihalcea, Rada and Pedersen, Ted},
  booktitle={Proceedings of the HLT-NAACL 2003 Workshop on Building and using parallel texts: data driven machine translation and beyond},
  pages={1--10},
  year={2003}
}

@inproceedings{vilar2006aer,
  title={AER: Do we need to “improve” our alignments?},
  author={Vilar, David and Popovi{\'c}, Maja and Ney, Hermann},
  booktitle={Proceedings of the Third International Workshop on Spoken Language Translation: Papers},
  year={2006}
}

@inproceedings{martelli2023xl,
  title={XL-WA: a Gold Evaluation Benchmark for Word Alignment in 14 Language Pairs},
  author={Martelli, Federico and Bejgu, Andrei Stefan and Campagnano, Cesare and {\v{C}}ibej, Jaka and Costa, Rute and Gantar, Apolonija and Kallas, Jelena and Koeva, Svetla Peneva and Koppel, Kristina and Krek, Simon and others},
  booktitle={Proceedings of the 9th Italian Conference on Computational Linguistics (CLiC-it 2023)},
  pages={272--280},
  year={2023}
}

@article{volk2010challenges,
  title={Challenges in building a multilingual alpine heritage corpus},
  author={Volk, Martin and Bubenhofer, Noah and Althaus, Adrian and Bangerter, Maya and Furrer, Lenz and Ruef, Beni},
  year={2010},
  publisher={University of Zurich}
}

@inproceedings{dyer2013simple,
  title={A simple, fast, and effective reparameterization of IBM model 2},
  author={Dyer, Chris and Chahuneau, Victor and Smith, Noah A},
  booktitle={Proceedings of the 2013 conference of the North American chapter of the association for computational linguistics: human language technologies},
  pages={644--648},
  year={2013}
}

@article{och2003systematic,
  title={A systematic comparison of various statistical alignment models},
  author={Och, Franz Josef and Ney, Hermann},
  journal={Computational linguistics},
  volume={29},
  number={1},
  pages={19--51},
  year={2003},
  publisher={MIT Press One Rogers Street, Cambridge, MA 02142-1209, USA journals-info~…}
}

@article{gale1993program,
  title={A program for aligning sentences in bilingual corpora},
  author={Gale, William A and Church, Kenneth},
  journal={Computational linguistics},
  volume={19},
  number={1},
  pages={75--102},
  year={1993}
}

@article{chen2023extending,
  title={Extending context window of large language models via positional interpolation},
  author={Chen, Shouyuan and Wong, Sherman and Chen, Liangjian and Tian, Yuandong},
  journal={arXiv preprint arXiv:2306.15595},
  year={2023}
}

@article{peng2023yarn,
  title={Yarn: Efficient context window extension of large language models},
  author={Peng, Bowen and Quesnelle, Jeffrey and Fan, Honglu and Shippole, Enrico},
  journal={arXiv preprint arXiv:2309.00071},
  year={2023}
}
\end{document}